\documentclass{article}
\usepackage{ijcai22}

\usepackage{times}

\usepackage{soul}
\usepackage{url}
\usepackage[hidelinks]{hyperref}
\usepackage[utf8]{inputenc}
\usepackage[small]{caption}
\usepackage{graphicx}
\usepackage{amsmath}
\usepackage{booktabs}
\usepackage{overpic}
\usepackage{tabularx}
\usepackage{graphicx}
\usepackage{varwidth}
\usepackage{capt-of}%
\usepackage{booktabs}       
\usepackage{amsmath}
\usepackage{amssymb}
\usepackage{amsfonts}       

\usepackage[table]{xcolor}
\definecolor{mygray}{gray}{.92}

\usepackage{multirow}

\makeatletter
\newcommand{\thickhline}{%
    \noalign {\ifnum 0=`}\fi \hrule height 1pt
    \futurelet \reserved@a \@xhline
}

\title{RePFormer: Refinement Pyramid Transformer for Robust \\Facial Landmark Detection}

\author{
Jinpeng Li$^1$\and
Haibo Jin$^2$\and
Shengcai Liao$^3$\footnote{Shengcai Liao is the corresponding author.}\and
Ling Shao$^4$\And
Pheng-Ann Heng$^{1,5}$
\\
\affiliations
$^1$Department of Computer Science and Engineering,\\
The Chinese University of Hong Kong, Hong Kong, China\\
$^2$Department of Computer Science and Engineering,\\ The Hong Kong University of Science and Technology, Hong Kong, China\\
$^3$Inception Institute of Artificial Intelligence (IIAI), UAE\\
$^4$Terminus Group, China\\
$^5$Guangdong Provincial Key Laboratory of Computer Vision and Virtual Reality Technology,\\ 
Shenzhen Institutes of Advanced Technology, Chinese Academy of Sciences, Shenzhen, China\\
\emails
jpli21@cse.cuhk.edu.hk,
haibo.nick.jin@gmail.com,
\{scliao, ling.shao\}@ieee.org,
pheng@cse.cuhk.edu.hk
}
\begin{document}

\maketitle
\begin{abstract}
This paper presents a Refinement Pyramid Transformer (RePFormer) for robust facial landmark detection. Most facial landmark detectors focus on learning  representative image features. However, these CNN-based feature representations are not robust enough to handle complex real-world scenarios due to ignoring the internal structure of landmarks, as well as the relations between landmarks and context. In this work, we formulate the facial landmark detection task as refining landmark queries along pyramid memories. Specifically, a pyramid transformer head (PTH) is introduced to build both homologous relations among landmarks and heterologous relations between landmarks and cross-scale contexts.
Besides, a dynamic landmark refinement (DLR) module is designed to decompose the landmark regression into an end-to-end refinement procedure, where the dynamically aggregated queries are transformed to residual coordinates predictions. Extensive experimental results on four facial landmark detection benchmarks and their various subsets demonstrate the superior performance and high robustness of our framework.
\end{abstract}

\section{Introduction}
\begin{figure}[t]
    \centering
    \includegraphics[width=\linewidth]{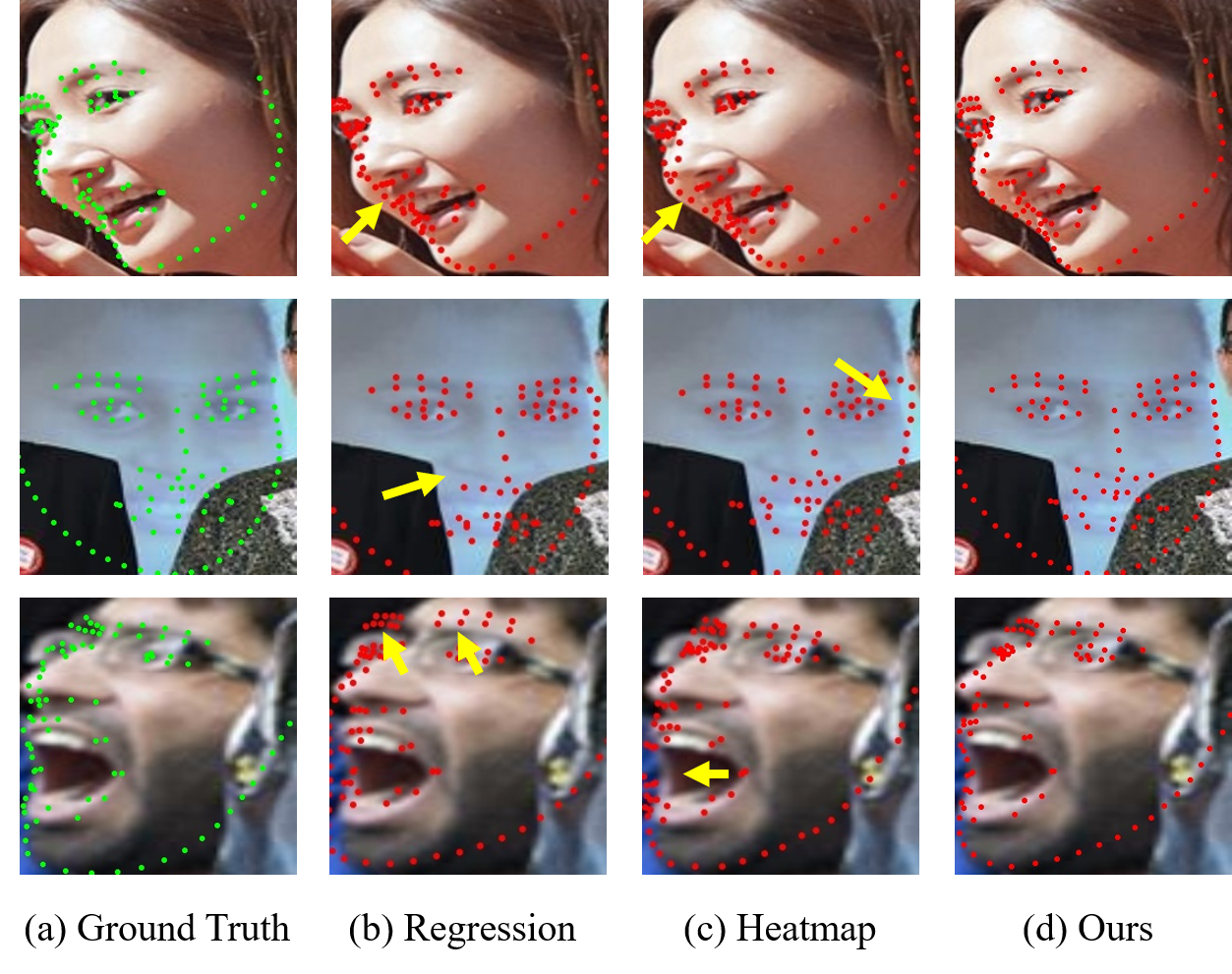}
    \vspace*{-0.2cm}
    \caption{Facial landmark detection results on WFLW. Yellow arrows point to the wrong results. (a) Ground truth. (b) Results of regression-based method. (c) Results of heatmap-based method. (d) Our results. Methods (b) and (c) ignore the internal structures of landmarks, while our method explores the internal structure of landmarks and builds the relations between landmarks and cross-scale contexts, thereby achieving more robust detection results.}
    \vspace*{-0.2cm}
    \label{fig:temp}
\end{figure}

Facial landmark detection aims to localize a set of predefined key points on 2D facial images. This task has attracted significant attention due to its wide range of applications~\cite{DBLP:journals/sensors/Ko18}. With the remarkable success of convolutional neural networks (CNNs), modern facial landmark detectors~\cite{WSC19,QSW19,JLS20} have achieved encouraging performance in constrained environments. However, as fundamental components in real-world facial applications, facial landmark detectors need to be able to consistently generate robust results for complex situations, which remains a significant challenge for existing methods. Figure~\ref{fig:temp} illustrates some examples of challenging face images and the detection results of existing algorithms and our method.

According to the ways of generating landmark coordinates from feature maps, previous works can be generally classified into three categories: heatmap-based methods, regression-based methods, and hybrid methods. Heatmap-based methods~\cite{WSC19,DoY19} treat facial landmark detection as a segmentation task, where pixels belong to the classes of landmarks and background. These methods mainly face two difficulties: 1) To reduce the quantization error in CNN backbones, they usually leverage single-scale high-resolution heatmaps~\cite{RFB15} to represent the surrogate results of landmarks, which introduces high computational costs and prevents them from fully exploring the pyramid features. 2) They lack global constraints on facial landmarks and ignore the relationships between image contexts and landmarks, which decreases their robustness to complex scenes. Regression-based methods~\cite{LSX17} directly transform image features into landmark coordinates with high efficiency. However, a single regression step is difficult to achieve satisfactory performance. Cascaded regression steps~\cite{LSX17,FKA18} are leveraged to refine results, but they are not fully end-to-end trainable and their performance is prone to saturation. Recently, Sun et al.~\cite{DBLP:conf/eccv/SunXWLW18} combined heatmap and regression together to inherit the advantages of both. However, their network designs are imbalanced, with most of the computational costs concentrated on the backbone. Tan et al.~\cite{DBLP:conf/cvpr/TanPL20} demonstrated that imbalanced network architectures can limit the detection performance. Moreover, existing detectors often focus on directly learning the landmark representation from pure image features, and ignore the dynamics and interactivity of landmarks and context information. \\

To address above limitations, we propose a novel facial landmark detector named the Refinement Pyramid Transformer (RePFormer) which focuses on robust detection for complex scenarios. We treat facial landmarks as attention quires, and formulate the facial landmark detection as a task of refining landmark queries through pyramid memories in an end-to-end trainable procedure. Specifically, we propose a pyramid transformer head (PTH) to mimic the zooming-in mechanism. In PTH, cross-scale attention constructs pyramid memories to combine long-range context and propagate high-level semantic information into lower levels. Then, landmark-to-landmark and landmark-to-memory attentions are employed in each PTH stage to enable dynamic interactions of landmarks and pyramid memories, which helps to regress the landmarks in various scenarios. Besides, we design a dynamic landmark refinement (DLR) method to refine the landmark queries by predicting residual coordinates and dynamically aggregating queries in an end-to-end trainable way. The residual coordinates prediction decomposes the regression into multiple steps, and each step refines the landmark queries based on a specific level in pyramid memories, which fully leverages the multi-level information in pyramid memories and decreases the difficulty of prediction. We conduct extensive experiments on several facial landmark detection benchmarks. Our models achieve state-of-the-art performance and show strong robustness on complex scenarios.

\section{Related Work}
\textbf{Facial Landmark Detection.}
Early facial landmark detection methods mainly focus on deforming a statistical facial landmark model into a 2D image by an optimization procedure. These methods typically apply different constraints on the statistical models, such as object shapes and texture features~\cite{DBLP:journals/ijcv/WuJ19}. However, they are not robust under various scenarios in the wild. Recently, deep neural network (DNN) based methods~\cite{WSC19,JLS20} show promising performance on this task. Their architectures are usually composed of a CNN backbone and a variant of a heatmap-based or/and regression-based detection head. The pixel values of heatmaps represent the likelihood of landmarks exiting in the corresponding positions~\cite{WSC19,DoY19}. A drawback of them is that the down-sampling ratio of heatmaps introduces quantization error. Thus, high-resolution networks, such as U-Net~\cite{RFB15} and HRNet~\cite{WSC19}, are commonly applied to address this difficulty. Instead of utilizing heatmaps as surrogate
results, regression-based methods~\cite{LSX17} directly predict facial landmarks by transforming image features into 2D coordinates. Cascaded regression steps~\cite{LSX17,FKA18} are widely applied to further refine the predicted coordinates. They usually follow an iterative cropping-and-regression pipeline which first crops patches from CNN feature maps and then sends them to carefully designed regressors, such as Recurrent Neural Network (RNN)~\cite{TSN16} or Graph Convolutional Network (GCN)~\cite{LLZ20}. However, extracting patches limits the long-range information exchange and may hinder the end-to-end training due to the non-differentiable cropping operators.

\vspace{+0.5mm}
\noindent \textbf{Vision Transformer.}
Transformer~\cite{DBLP:conf/nips/VaswaniSPUJGKP17} is first proposed for sequence-to-sequence tasks in natural language processing. Its network architecture simply consists of self-attention based encoders and decoders to model dependencies among any positions in a sequence. Recently, computer vision, from high-level tasks to low-level tasks, has also benefited from its strong representation capability and long-range interactions~\cite{DBLP:conf/eccv/CarionMSUKZ20}. The Vision Transformer (ViT)~\cite{DBLP:journals/corr/abs-2010-11929} directly flattens image patches and processes their feature and position embeddings using a pure transformer encoder for the image classification task.  DETR~\cite{DBLP:conf/eccv/CarionMSUKZ20} introduces a transformer-based detector to solve the object detection problem as unordered set prediction. Its fully end-to-end design removes the hand-crafted components, such as anchor assignment and duplication reduction, in modern object detectors. Transformers have also been explored for landmark detection tasks. Yang et al.~\cite{DBLP:journals/corr/abs-2012-14214} leverage a transformer as a non-local module to build long-range spatial dependencies in image features for more explainable human pose estimation. In contrast, the transformer in our model explicitly establishes the relations between landmarks to landmarks, and landmarks to image contexts, and gradually refines the landmark queries along the pyramid transformer memories.

\begin{figure*}[t]
  \centering
  \includegraphics[width=0.85\linewidth]{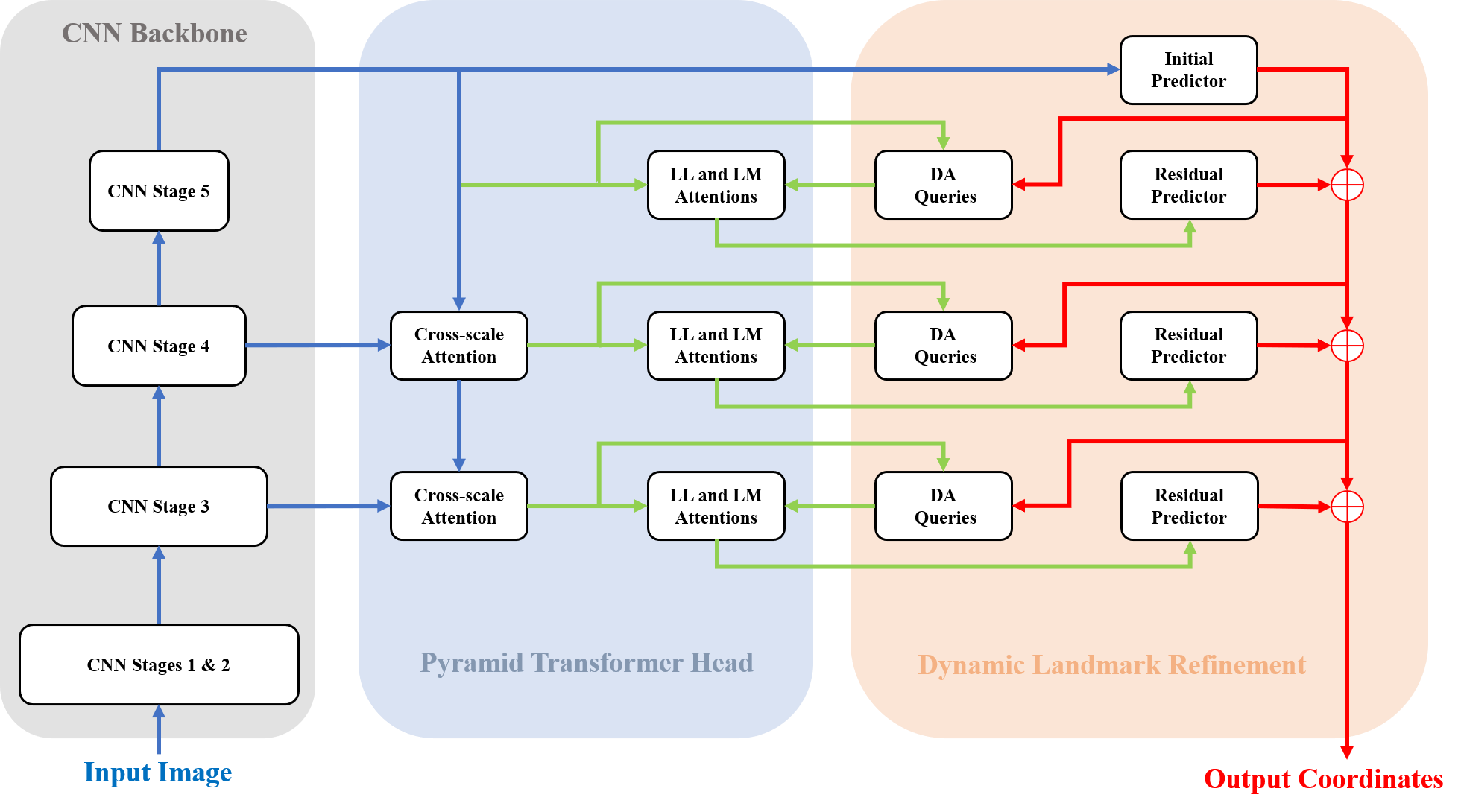}
  \caption{Overview of RePFormer. LL and LM are the landmark-to-landmark attention and landmark-to-memory attention, respectively. DA queries are dynamically aggregated queries. Blue lines represent feature maps. Green lines are the embeddings of memories and landmark queries. Red lines represent coordinates and residual predictions of facial landmarks.}
  \vspace*{-0.0cm}
  \label{fig:arch}
\end{figure*}

\section{Method}

\subsection{Overview}
\textbf{Problem Formulation.} Given an input 2D image $I$, our RePFormer aims at learning a facial landmark detector $f_\theta(I)$ to predict an ordered set $L \in \mathbb{R}^{N \times D}= \{l_1,...,l_N\}$ which represents $N$ ordered facial landmarks, where each landmark $l_i$ is in $D$-dimensional coordinates. Note that the transformer-based object detector, DETR~\cite{DBLP:conf/eccv/CarionMSUKZ20}, also formulates general object detection as a set prediction task. However, the predicted set of DETR is unordered and with variable length, since the sequence and number of objects are not fixed in general object detection datasets. In contrast, the facial landmark detection task assumes that each image only contains one main face (if there are several faces in an image, only the largest central one is considered), which is annotated with a fixed number of landmarks. Besides, instead of solving a bipartite matching as in DETR, the label assignment in our framework is simply defined as a fixed one-to-one way.

\noindent \textbf{Overall Architecture.}
The architecture of our RePFormer is illustrated in Fig.~\ref{fig:arch}. It is
built upon a CNN-based feature extractor and a transformer-based detection head. First, the input image is fed into a common CNN backbone which contains several network stages. The feature maps generated by these stages are of various resolutions and semantics, forming hierarchical image features. Next, our PTH module applies cross-scale attentions on these hierarchical feature maps to produce pyramid memories with multi-level semantic information. Then, the PTH stages use landmark-to-landmark and landmark-to-memory attentions to fuse the context information of memories into landmark queries and model the relations among landmarks. After each PTH stage, our DLR module performs mutual updates by predicting the residual coordinates of landmarks based on the status of current queries and evolving the landmark queries based on the coordinates of current landmarks. Thus, the whole facial landmark detection task is formulated as a step-by-step landmark refinement process.

\begin{figure*}[t]
  \centering
  \begin{overpic}[width=0.9\textwidth]{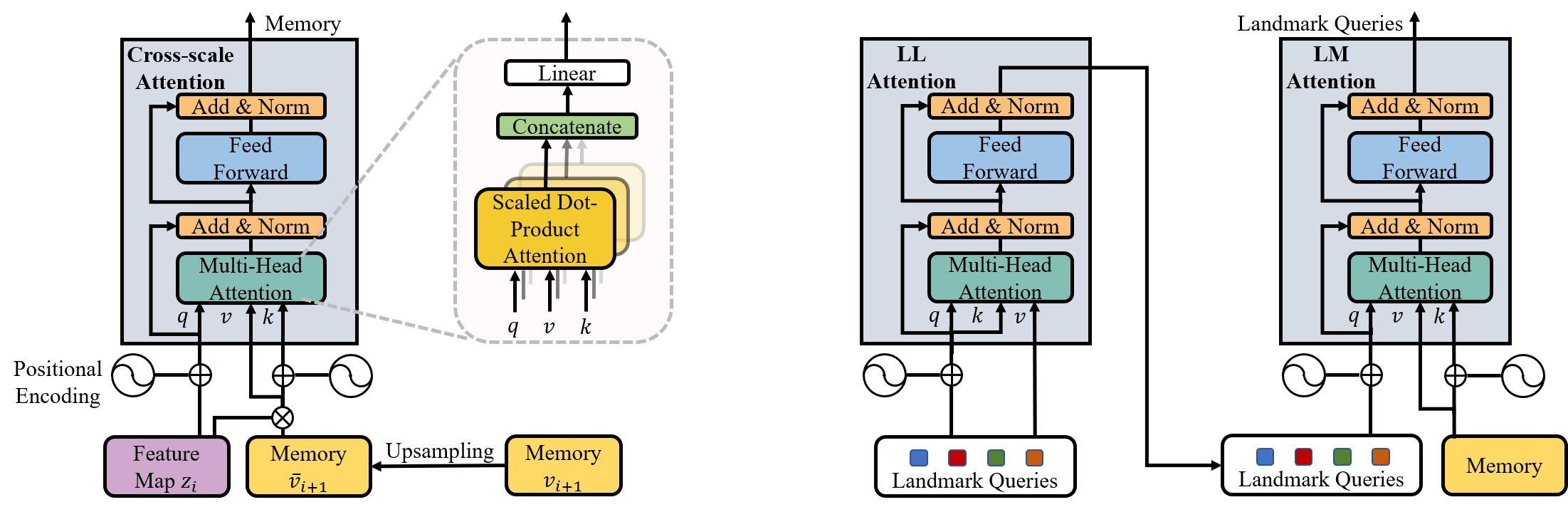}
    \put(18,-3){(a)}
    \put(73,-3){(b)}
    
    \end{overpic}
    \vspace{0.5cm}
  
  \caption{(a) Architecture of Cross-scale attention.  ~~~(b) Architectures and data flow of LL and LM attentions.}
  \vspace{-0.5cm}
  \label{fig:ll}
\end{figure*}

\subsection{Pyramid Transformer Head}

\noindent \textbf{Cross-Scale Attention.}
Feature pyramids have been explored for solving various challenges in computer vision, such as detecting objects at various scales~\cite{DBLP:conf/ispa/LiLWCYCLZ19} and combining different levels of semantic features for segmentation~\cite{RFB15}. In this work, our transformer-based detection head, PTH, explicitly employs pyramid features to obtain cross-scale semantic information, mimicking the zooming-in mechanism of human annotators. 
Our cross-scale attention takes the hierarchical feature set $Z$ as input, and generates the pyramid memory set $V$ following a top-down pathway. $Z$ is obtained by a CNN backbone $f_B$ which is defined as follows:
\begin{equation}
    f_B(I)=f_B^M(...f_B^i(...f_B^1(I))),
\end{equation}
where $M$ is the number of network stages, and $f_B^i$ is the $i$th network stage to take the feature maps $z_{i-1}$ as input and generate feature maps $z_i$ as follows:
\begin{equation}
    z_i=f_B^i(z_{i-1}).
\end{equation}
These hierarchical features form the CNN feature set $Z=\{z_M,..., z_1\}$, which contains multi-level feature maps with various resolutions and semantics. 
Top-level features contain strong semantics but less spatial information, making them only suitable for robustly locating the rough positions of landmarks. To better propagate high-level semantics into all levels of features and combine long-range information, our PTH computes the $i$th memory $v_i$ by applying a cross-scale attention on every two adjacent levels of feature maps $z_{i}$ and memories $v_{i+1}$, which is inspired by the top-down Grounding Transformer in~\cite{DBLP:conf/eccv/ZhangZTWHS20}. The architecture of cross-scale attention is shown in Fig.~\ref{fig:ll}(a), which takes three inputs including query $q_i$, key $k_i$ and value $\hat{v}_i$. Since the attention operation is permutation-invariant to input elements, $q_i$ is composed of $z_i$ and its fixed positional encodings $p_i$ by $q_i=z_i+p_i$ to maintain relative positions of pixels. $k_i$ and $\hat{v}_i$ are both obtained by fusing the information of $z_i$ and $v_{i+1}$. A bilinear interpolation upsamples $v_{i+1}$ into $\bar{v}_{i+1}$ with the same resolution as $z_i$. Then $\hat{v}_i$ is generated by a Hadamard product as $\hat{v}_i=z_i\circ\bar{v}_{i+1}$, where $\bar{v}_{i+1}$ aims to mask output the noisy low-level details outside semantic areas and maintains the precise spatial information within semantic areas. To make the similarity calculation between $q_i$ and $k_i$ semantic and position sensitive, $k_i$ is also supplemented with fixed positional encodings as $k_i=\hat{v}_i+p_i$. In summary, the memory ${v_i}$ is computed as

\begin{equation}
    \hat{z_i}=\text{LN}(z_i+\text{MHA}(q_i, k_i, \hat{v}_i))
\end{equation}

\begin{equation}
\begin{cases}
    v_i={\text{LN}}(\hat{z_i}+{\text{FFN}}(\hat{z_i})),& \text{if } i<M \\
    v_i=z_i,              & \text{otherwise},
\end{cases}
\end{equation}
where MHA is a multi-head attention~\cite{DBLP:conf/nips/VaswaniSPUJGKP17}, FFN is a feed forward network, and LN is a layer normalization. Thus, the pyramid memories form the set $V=\{v_M, ...,v_1\}$, where each $v_i$ has the same resolution as $z_i$ but with cross-level and long-range information.

\noindent \textbf{PTH Forwarding.}
Our PTH detection head $f_H$ is composed of cascaded transformer stages, and defined as follows:
\begin{equation}
    f_H(V, E^1)=f_{H}^K(v_{M-K+1},...f_H^i(v_{M-i+1},...f_H^1(v_{M}, E^1))),
\end{equation}
where $K$ is the number of transformer stages and $f_H^i$ is the $i$th stage. PTH introduces the landmark query set $E\in \mathbb{R}^{N \times D}=\{e_1,...,e_N\}$ to store the states of landmarks, the size of which is the same as the number of landmarks. $E$ can either be initialized from trainable embeddings as in DETR~\cite{DBLP:conf/eccv/CarionMSUKZ20}, or dynamiclly aggregated from memories as in our method, which will be introduced in next section. PTH takes the pyramid and initial landmark query set $E^1$ as inputs and repeatedly updates the states of $E$ based on the information from pyramid memories $V$, in a top-down pathway. This enables it to first robustly locate the rough positions of landmarks in the top-level memory, and then gradually integrates more and more precise spatial information from lower levels of memories. Specifically, $f_H^i$ is the $i$th PTH stage to take the memory $v_{M-i+1}$ and the landmark query set $E^i$ in the $i$th state as inputs and generate the new query state $E^{i+1}$ for the next PTH stage. This is computed by
\begin{equation}
   \begin{split}
    E^{i+1}&=f_H^i(v_{M-i+1}, E^i) \\
    &=g_{LM}(g_{LL}(E^i), v_{M-i+1})
\end{split}
\end{equation}
where each PTH stage is composed of two successive attentions, i.e. a landmark-to-landmark attention $g_{LL}$ and a landmark-to-memory attention $g_{LM}$ as shown in Fig.~\ref{fig:ll}(b), which is inspired by the transformer decoders in~\cite{DBLP:conf/nips/VaswaniSPUJGKP17,DBLP:conf/eccv/CarionMSUKZ20}. They are both implemented with multi-head attentions. The query, key and value of $g_{LL}$ all come from $E^i$, and the query, key and value of $g_{LM}$ are the output of $g_{LL}$, $v_{M-i+1}$ and $v_{M-i+1}$, respectively. $g_{LL}$ models the dynamic relations among landmark queries using their similarities, and updates their status. $g_{LM}$ computes the relations among landmarks and pyramid memories to endow cross-scale information and long-range image context into landmark queries. These dynamic relations are built on the fly and specific to each image, thus they are more robust in detecting facial landmarks without salient visual features.

\begin{table*}
\footnotesize
\centering

\newcolumntype{C}{>{\centering\arraybackslash}X}%
\begin{tabularx}{\textwidth}{llCCCCCCCCCCC}
\hline\thickhline
\rowcolor{mygray} 	
 & &\multicolumn{7}{c}{WFLW} & & &  \\ \cline{3-9} 
\rowcolor{mygray} 
\multirow{-2}{*}{Method}  & \multirow{-2}{*}{Backbone} & Pose  & Expr. & Illu. & M.u. & Occ. & Blur & Full & \multirow{-2}{*}{300W} & \multirow{-2}{*}{COFW} & \multirow{-2}{*}{AFLW}\\
\hline \hline	
DVLN ~\cite{WuY17}  & VGG-16 & 11.54 & 6.78 & 5.73 & 5.98 & 7.33 & 6.88 & 6.08 & 4.66 & - & - \\
LAB ~\cite{WQY18}     & ResNet-18 & 10.24 & 5.51 & 5.23 & 5.15 & 6.79 & 6.32 & 5.27 & 3.49 & 5.58 & 1.85 \\
Wing~\cite{FKA18}     & ResNet-50 & 8.43  & 5.21 & 4.88 & 5.26 & 6.21 & 5.81 & 4.99 & - & 5.07 & 1.47\\
DeCaFA~\cite{DBC19}  & Cascaded U-net & - & - & - & - & - & - & 5.01 & 3.69 & - & -\\ 
HRNet~\cite{WSC19}    & HRNetV2-W18 & 7.94  & 4.85 & 4.55 & 4.29 & 5.44 & 5.42 & 4.60 & 3.32   & 3.45   & 1.57\\
AVS~\cite{QSW19}   & ResNet-18 & 9.10 & 5.83 & 4.93 & 5.47 & 6.26 & 5.86 & 5.25 & 4.54 & - & -\\ 
AVS w/ LAB~\cite{QSW19}   & Hourglass & 8.21 & 5.14 & 4.51 & 5.00 & 5.76 & 5.43 & 4.76 & 4.83 & - & - \\ 
AVS w/ SAN ~\cite{QSW19}   & - & 8.42 & 4.68 & 4.24 & 4.37 & 5.60 & 4.86 & 4.39 & 3.86 & - & - \\ 

AWing~\cite{WBF19}    & Hourglass & 7.38 & 4.58 & 4.32 & 4.27 & 5.19 & 4.96 & 4.36  & 3.07 & - & -\\
STYLE~\cite{QSW19}  & ResNet-18 & 8.42 & 4.68 & 4.24 & 4.37 & 5.60 & 4.86 & 4.39 & 4.54 & - & -\\

LUVLi~\cite{KMM20}   & DU-Net & - & - & - & - & - & - & 4.37 & 3.23 & - & -\\ 
DAG~\cite{LLZ20}  & HRNet-W18 & 7.36 & 4.49 & 4.12 & 4.05 & \textcolor{red}{\textbf{4.98}} & \textcolor{blue}{\textbf{4.82}} & 4.21 & 3.04 & - & -\\
PIPNet ~\cite{JLS20} & ResNet-18 & 8.02 & 4.73 & 4.39 & 4.38 & 5.66 & 5.25 & 4.57 & 3.36   & 3.31   & 1.48\\
PIPNet ~\cite{JLS20}  & ResNet-50 & 7.98 & 4.54 & 4.35 & 4.27 & 5.65 & 5.19 & 4.48 & 3.24   & 3.18   & 1.44\\
PIPNet ~\cite{JLS20}  & ResNet-101 & 7.51 & 4.44 & 4.19 & 4.02 & 5.36 & 5.02 & 4.31 & 3.19   & 3.08   & \textcolor{red}{\textbf{1.42}}\\
\hline
RePFormer (ours)     & ResNet-18  & 7.38 & 4.28 & \textcolor{blue}{\textbf{4.06}} &	4.04 & 5.17 & 4.86 & 4.20 & 3.07   & 3.07   & 1.44\\ 
RePFormer (ours)     & ResNet-50  & \textcolor{blue}{\textbf{7.31}} & \textcolor{blue}{\textbf{4.25}} & 4.09 &	\textcolor{blue}{\textbf{3.94}} & 5.15 & \textcolor{blue}{\textbf{4.82}} & \textcolor{blue}{\textbf{4.14}} & \textcolor{blue}{\textbf{3.03}}   & \textcolor{red}{\textbf{3.01}} & \textcolor{blue}{\textbf{1.43}}\\
RePFormer (ours)    & ResNet-101  & \textcolor{red}{\textbf{7.25}} & \textcolor{red}{\textbf{4.22}} & \textcolor{red}{\textbf{4.04}} & \textcolor{red}{\textbf{3.91}} & \textcolor{blue}{\textbf{5.11}} & \textcolor{red}{\textbf{4.76}} & \textcolor{red}{\textbf{4.11}} & \textcolor{red}{\textbf{3.01}}   & \textcolor{blue}{\textbf{3.02}}   & \textcolor{blue}{\textbf{1.43}}\\
\hline
\end{tabularx}
\caption{Benchmarking results of state-of-the-art methods and our models on the WFLW including the full set and six subsets, 300W, COFW and AFLW datasets. The best and second best results are marked in colors of \textcolor{red}{red} and \textcolor{blue}{blue}, respectively.}
\vspace*{-0.3cm}
\label{tab:wflw}
\end{table*}

\subsection{Dynamic Landmark Refinement}
Our model directly predicts the coordinates of landmarks, so it can be regarded as a regression-based method. Compared to heatmap-based detectors, a single regression step is hard to generate competitive results. Thus, multiple cropping-then-detection steps~\cite{TSN16} are used to improve the performance of regression-based methods in a coarse-to-fine way. However, existing multi-step methods usually have two drawbacks: 1) The hard cropping operator is not differentiable with respect to the input coordinates. 2) The detection is performed only based on the cropped features, without access to the image context information. To address these challenges, our RePformer introduces the DLR module to modulate facial landmark detection into a fully end-to-end refinement process using residual coordinates predictions and dynamically aggregated queries. 

\vspace{+1mm}
\noindent \textbf{Residual Coordinates Prediction.} The initial facial landmark set $L^0$ is predicted based on the highest-level memory $v_M$. Then, our DLR appends additional predictors to higher PTH stages, and the $i$th predictor $f_P^i$ only needs to predict the residual coordinates of landmarks $U^i \in \mathbb R^{N \times D}= \{u_1^i,...,u_N^i\}$ with respect to $L^{i-1}$, as follows:
\begin{equation}
    u_j^i=f_P^i(e_j^i),
\end{equation}
where $u_j^i$ are the residual coordinates of the $j$th landmark in the $i$th stage, and $e_j^i$ is the $j$th landmark query in the $i$th stage. The predictor $f_P^i$ is a two-layer FFN which is agnostic to different landmarks, because it only needs to predict the residual values instead of the absolute coordinates. And the $i$th landmark set $L^i$ is computed by:
\begin{equation}
    L^i=\{l_1^{i-1}+u_1^i,...,l_N^{i-1}+u_N^i\}.
\end{equation}
L1 loss is used as the loss function between the ground-truths and $L^i$ generated by the $i$th PTH stage. Thus, with this step-by-step refinement, our DLR gradually pushes the regressed landmarks closer to the ground-truth coordinates. 

\vspace{+1mm}
\noindent \textbf{Dynamically Aggregated Queries.} To adapt the transformer architecture to the residual prediction task, the landmark query set $E^i$, as the inputs of the $i$th PTH stage, need to represent the status of the current results $L^{i-1}$ using their semantic and spatial information. To solve this challenge in an end-to-end way, a dynamic aggregation method is proposed to extract landmark queries by aggregating pyramid memories weighted by their relative positional information. This can be seen as a soft version of the "cropping" operator, and the $j$th query for stage $i$ is computed by as follows
\begin{equation}
    e_j^i=\sum_{k\in \Omega}s_{ijk}\cdot v_{M-i+1}^k,
\end{equation}
where $k$ is a pixel's memory index, $\Omega$ is the domain of all indexes, and $v_{M-i+1}^k$ is the $k$th memory pixel in the stage $M-i+1$. The normalized similarity $s_{ijk}$ between query $e_j^{i-1}$ and memory pixel $v_{M-i+1}^k$ is represented by
\begin{equation}
    s_{ijk}=\frac{exp({-\lVert l_j^{i-1}-c_k \rVert}\cdot\tau)}{\sum_{\hat{k}\in \Omega}exp({-\lVert l_j^{i-1}-c_{\hat{k}}\rVert}\cdot\tau)},
\end{equation}
where $\lVert . \rVert$ is the squared L2 norm, and $c_k$ represents the coordinates of memory pixel $v_{M-i+1}^k$. $\tau$ is a temperature parameter which is bigger than 1 to amplify the differences in coordinates. Compared with hard cropping, dynamically aggregated queries are differentiable to coordinates of landmarks, making the entire multi-step refinement procedure fully end-to-end trainable. Besides, these queries are composed of the whole memory to explore both the position sensitive semantic features and global image context information.

\section{Experiments}
\subsection{Implementation Details and Datasets}
We use ResNet~\cite{DBLP:conf/cvpr/HeZRS16} pretrained on ImageNet~\cite{DBLP:conf/cvpr/DengDSLL009} as the backbone of RePFormer, and the default depth of backbone is 18 unless otherwise specified. The Adam optimizer without weight decay is used to train our models, $\beta1$ and $\beta2$ are set to 0.9 and 0.999, respectively. All models are trained for 360 epochs with a batch size of 16. The initial learning rate is 0.0001, which is decayed by a factor of 10 after 200 epochs. The temperature $\tau$ is set to 1000. The L1 loss is used as the loss function for all outputs, and the loss weights are simply set to 1. We conduct experiments on four popular facial landmark detection datasets including WFLW~\cite{WQY18}, 300W~\cite{STZ13}, AFLW-Full~\cite{KWR11}, and COFW~\cite{BPD13}. Most of our settings follow PIPNet~\cite{JLS20}. All input images are resized to 256x256. 

\subsection{Comparison with State-of-the-art Approaches}
We compare RePFormer with the state-of-the-art facial landmark detection methods using the evaluation metric of normalized mean error (NME). 

\vspace{+1mm}
\noindent \textbf{WFLW.}
Table~\ref{tab:wflw} shows the performance of state-of-the-art methods and our RePFormer models with three backbones including ResNet-18, ResNet-50 and ResNet-101. Thanks to our effective detection head, on the full set of WFLW, our model with the lightweight ResNet-18 backbone already achieves better performance (4.20\% NME) than all existing methods including those with much heavier backbones, such as Cascaded U-net and Hourglass. With ResNet-50 and ResNet-101, RePFormer further improves results to 4.14\% and 4.11\% NME, outperforming the most competitive method, DAG~\cite{LLZ20}, by 1.7\% and 2.4\%, respectively. We also conduct experiments on six subsets of WFLW. Our models outperform all previous methods on five subsets, and achieve all the best and second best results, demonstrating the strong performance and robustness of our framework in various evaluation scenarios.

\vspace{+1mm}
\noindent \textbf{300W.}
The third-to-last column of Table~\ref{tab:wflw} compares the performance of our models and state-of-the-art methods on the full set of 300W. All the reported results are normalized by the inter-ocular distances. Our RePFormers with ResNet-101 and ResNet-50 backbones achieve the best and second-best results, respectively. Only with the lightweight backbone of ResNet-18, our method already significantly outperforms most of the state-of-the-art methods, such as LAB~\cite{WQY18}, STYLE~\cite{QSW19}, HRNet~\cite{WSC19}, and PIPNet~\cite{JLS20}. Note that although RePFormer has a deeper detection head than detectors with specially tailored head for time efficiency, we argue that their architectures of large backbones with small detection heads are not optimal, and our RePFormer head can achieve better performance gain than using heavier backbone. Tan et al.~\cite{DBLP:conf/cvpr/TanPL20} also showed that balanced architectures can yield higher accuracy-speed ratio. For example, compared to the competitive method, PIPNet~\cite{JLS20}, our RePFormer with ResNet-18 outperforms PIPNet with ResNet-101 by 3.8\% with similar inference speed (56 FPS vs. 59 FPS).

\noindent \textbf{COFW.} We compare our methods with the state-of-the-art works on the COFW dataset under the intra-data setting which is shown in the second-to-last column of Table~\ref{tab:wflw}. The inter-ocular distances are used to normalize the results. Our models with three different backbones achieve the top-3 performances among all methods.

\noindent \textbf{AFLW.}
Comparative results of our models and state-of-the-art results are shown in the last column of Table~\ref{tab:wflw}. Following previous works, all landmark coordinates are normalized by the image size. As can be seen that our RePFormers with different backbones achieve the second-best and third-best results, while only PIPNet~\cite{JLS20} with ResNet-101 slightly outperforms us (1.42\% NME vs. 1.43\% NME).

\begin{table}[t]
    \centering
    
    \scalebox{0.95}{
    \begin{tabular}{ c c| r r r}
    \hline
    \thickhline
    \rowcolor{mygray}  
     \multicolumn{2}{c}{Modules}               &  \multicolumn{3}{c}{300W}\\\cline{1-2}  \cline{3-5} 
     \rowcolor{mygray}  
      \hline
      PTH& DLR & Full      & Common & Challenge                   \\ 
\hline \hline  
 &   & 3.28 & 2.92 & 4.99                         \\
$\surd$ &   &3.23 & 2.88 & 4.93                         \\
 & $\surd$  & 3.22& 2.92 & 4.85                         \\
$\surd$&  $\surd$ & \textbf{3.07} & \textbf{2.72} & \textbf{4.69}
                      \\
      \bottomrule
    \end{tabular}}
    \caption{Comparative results on 300W by using different components in RePFormer.}
    \label{tab:components}
\vspace*{-0.4cm}
\end{table}

\begin{table}[t]
    \centering
    
    \begin{tabular}{ c |c c c c}
    \hline 
    \thickhline
    \rowcolor{mygray} 
    $\tau$ & 10 & 100 & 1000 & 10000\\
    \hline \hline
    NME(\%) & 3.22 & 3.16 & \textbf{3.07} & 3.16\\
    \bottomrule
    
    \end{tabular}
    \caption{Performance of different $\tau$ on the full set of 300W.}
    \label{tab:tau}
\end{table}

\subsection{Ablation Studies}
\noindent \textbf{RePFormer Components.} Table~\ref{tab:components} shows the peformance of RePFormer with different componets. First, we construct a strong baseline model by appending a three-stage transformer detection head on a CNN backbone. Note that only the feature maps from the last stage of the CNN are used as the memory for the transformer, and landmark embeddings are fixed after training. Thanks to the long-range information modeling of transformer, our baseline model already achieves competitive performance compared to the state-of-the-art methods. Secondly, we apply PTH module on the baseline model. With the help of PTH, the performance of our model is improved on all sets, which demonstrates the effectiveness of pyramid memories with cross-scale information. Thirdly, we integrate DLR into our baseline model. From Table~\ref{tab:components}, we observe that DLR significantly improves the performance on the challenge subset. Finally, as shown in the last row of Table~\ref{tab:components}, our RePFormer with PTH and DLR modules achieves large performance improvements over the baseline on all subsets. Note that, compared to the baseline, the performance gain of RePFormer is even larger than the sum of improvements of all individual components, which demonstrates that the two proposed components are complementary to each other and can work together to promote more accurate regression.

\vspace{+1mm}
\noindent \textbf{Temperature Values.} We evaluate the performance of RePFormer with different temperature values on the full set of 300W. $\tau$ controls the weight map for dynamically aggregated queries. As shown in Table~\ref{tab:tau}, the performance is continuously improved as the $\tau$ value increases from 10 to 1000, but an excessively large $\tau$ value decreases the performance. The results indicate that dynamic landmark queries mainly aggregate information from nearby pixels, while long-range information is also valuable for generating accurate results.

\section{Conclusion}
In this paper, we present a refinement pyramid transformer, RePFormer, for facial landmark detection. The proposed PTH utilizes a cross-scale attention to generate pyramid memories containing multi-level semantic and spatial information. Landmark-to-landmark and landmark-to-memory attentions are employed in each PTH stage to fuse memory information into landmark queries and model long-range relationships between landmarks. A DLR module is introduced to solve the task of landmark regression by a fully end-to-end multi-step refinement procedure where the landmark queries are gradually refined by a dynamic aggregation method through top-down pyramid memories.

\section*{Acknowledgments}
This work is supported by Hong Kong Research Grants Council
under General Research Fund Project No. 14201620.

\bibliographystyle{named}
\bibliography{ijcai22}
\end{document}